\documentclass{article}
\usepackage{spconf,amsmath,graphicx}
\usepackage{url}
\usepackage{hyperref}
\usepackage{booktabs}
\usepackage{subcaption}
\usepackage[normalem]{ulem}
\useunder{\uline}{\ul}{}
\usepackage{float}
\usepackage{amsmath}
\usepackage{amsfonts}
\usepackage{multirow}
\usepackage{dsfont}

\title{Conditioned Query Generation for Task-Oriented Dialogue Systems}

\name{St\'{e}phane d'Ascoli$^{\diamond}$$^{\bullet}$ \qquad \hspace{-0.6cm} Alice Coucke\sthanks{Corresponding author: \texttt{alice.coucke@snips.ai}}$^{\diamond}$ \qquad \hspace{-0.6cm} Francesco Caltagirone$^{\diamond}$ \qquad \hspace{-0.6cm} Alexandre Caulier$^{\diamond}$ \qquad \hspace{-0.6cm} Marc Lelarge$^{\odot}$}
\address{$^{\diamond}$ Snips, Paris, France \\ 
$^{\bullet}$Laboratoire de Physique Statistique, École Normale Supérieure, \\PSL Research University, Paris, France \\
$^{\odot}$INRIA-ENS, Paris, France
}

\begin{document}
%\ninept
%
\maketitle
\thispagestyle{plain}
\pagestyle{plain}
\begin{abstract}

    Scarcity of training data for task-oriented dialogue systems is a well known problem that is usually tackled with costly and time-consuming manual data annotation. An alternative solution is to rely on automatic text generation which, although less accurate than human supervision, has the advantage of being cheap and fast.
    In this paper we propose a novel controlled data generation method that could be used as a training augmentation framework for closed-domain dialogue.
    Our contribution is twofold. First we show how to optimally train and control the generation of intent-specific sentences using a conditional variational autoencoder. Then we introduce a novel protocol called \textit{query transfer} that allows to leverage a broad, unlabelled dataset to extract relevant information. Comparison with two different baselines shows that our method, in the appropriate regime, consistently improves the diversity of the generated queries without compromising their quality\footnote{The PyTorch code for the experiments is publicly available on GitHub at \url{https://github.com/snipsco/automatic-data-generation}.}.

\end{abstract}

\section{Introduction}

    Closed-domain dialogue systems, single- or multi-turn, have become ubiquitous nowadays with the rise of conversational interfaces. These systems aim at extracting relevant information from a user's spoken query, produce the appropriate response/action and, when applicable, start a new dialogue turn. The typical spoken language understanding (SLU) framework relies on a speech-recognition engine that transforms the spoken utterance into text followed by a natural language understanding engine that extracts meaning from the text utterance. Here we consider essentially single-turn closed-domain dialogue systems where the meaning is well summarized by an intent and its corresponding slots. As an example, the query ``Play Skinny Love by Bon Iver'' should be interpreted as a {\it PlayTrack} intent with slots {\it TrackTitle} ``Skinny Love'' and {\it Artist} ``Bon Iver''.
    
    Training data for conversational systems consist in annotated utterances corresponding to the various intents within the scope of the system. When developing a new interaction scheme with new intents, a (possibly large) representative set of manually annotated utterances needs to be produced, which is a costly and time-consuming process. It is therefore desirable to automate it as much as possible to reduce cost and development time. We aim at alleviating the training data scarcity problem through automatic generation of utterances conditioned to the desired intent. 
    
    In this work, we focus on the conditioned generation problem in itself in the context of conversational systems and detail the assessment of its performance in terms of quality and diversity of generated sentences. We choose to consider the low data regime as we feel it is prevalent for this task. The application of the proposed approach as an augmentation technique for SLU tasks will be the subject of a future paper and is beyond the scope of this work.
    %Keeping in mind the final goal of data-augmentation for SLU training, in this work we will focus on the conditioned generation problem in itself and we will detail the analysis of its performance as an augmentation method in an upcoming paper.
    We consider the scenario in which only a small set of annotated queries is available for all the in-domain intents, while leveraging a very large reservoir of unannotated queries that belong to a broad spectrum of intents ranging from close to far domain. This situation is indeed very typical of conversational platforms like DialogFlow, IBM Watson, or Snips, which offer a high degree of user customization.
    
    %We train a Conditional Variational autoencoder (CVAE)~\cite{cvae_2015} on both the annotated in-domain and the unannotated out-of-domain data and tune the model hyper-parameters to favor transfer of valuable knowledge from the reservoir. We then use the trained CVAE decoder to generate new queries for each intent. We call this mechanism {\it query transfer}.\\ 
    
    \subsection{Contribution and Outline. }
        In this paper we propose a method for conditional text generation with Conditional Variational Autoencoders (CVAE)~\cite{cvae_2015}, that leverages transfer from a large out-of-domain unlabelled dataset. The model hyper-parameters are tuned to favor transfer of valuable knowledge from the reservoir while maintaining an accurate conditioning. We use the trained CVAE decoder to generate new queries for each intent. We call this mechanism {\it query transfer}. We analyse the performance of this approach on the publicly-available Snips dataset \cite{coucke2018snips} through both quality and diversity metrics. We also observe an improvement in the perplexity of a language model trained on data augmented with our generation scheme. This preliminary result is encouraging for future application to SLU data augmentation. We briefly show in the Appendix that the same approach can be applied to computer vision as well.
        
        The paper is structured as follows: in Section \ref{sec: rel_work}, we briefly present the related literature, in Section \ref{sec: approach} we introduce our approach in details, and in Section \ref{sec: experiments} we describe the experimental settings and the evaluation metrics. In Section \ref{sec: results} we show our results on the quality of generation compared to two different baselines and on language modelling perplexity, before concluding in Section \ref{sec: conclusions}.

\begin{figure*}[htb]
    \centering
    \begin{subfigure}{0.55\linewidth}
    \includegraphics[width=1.0\textwidth]{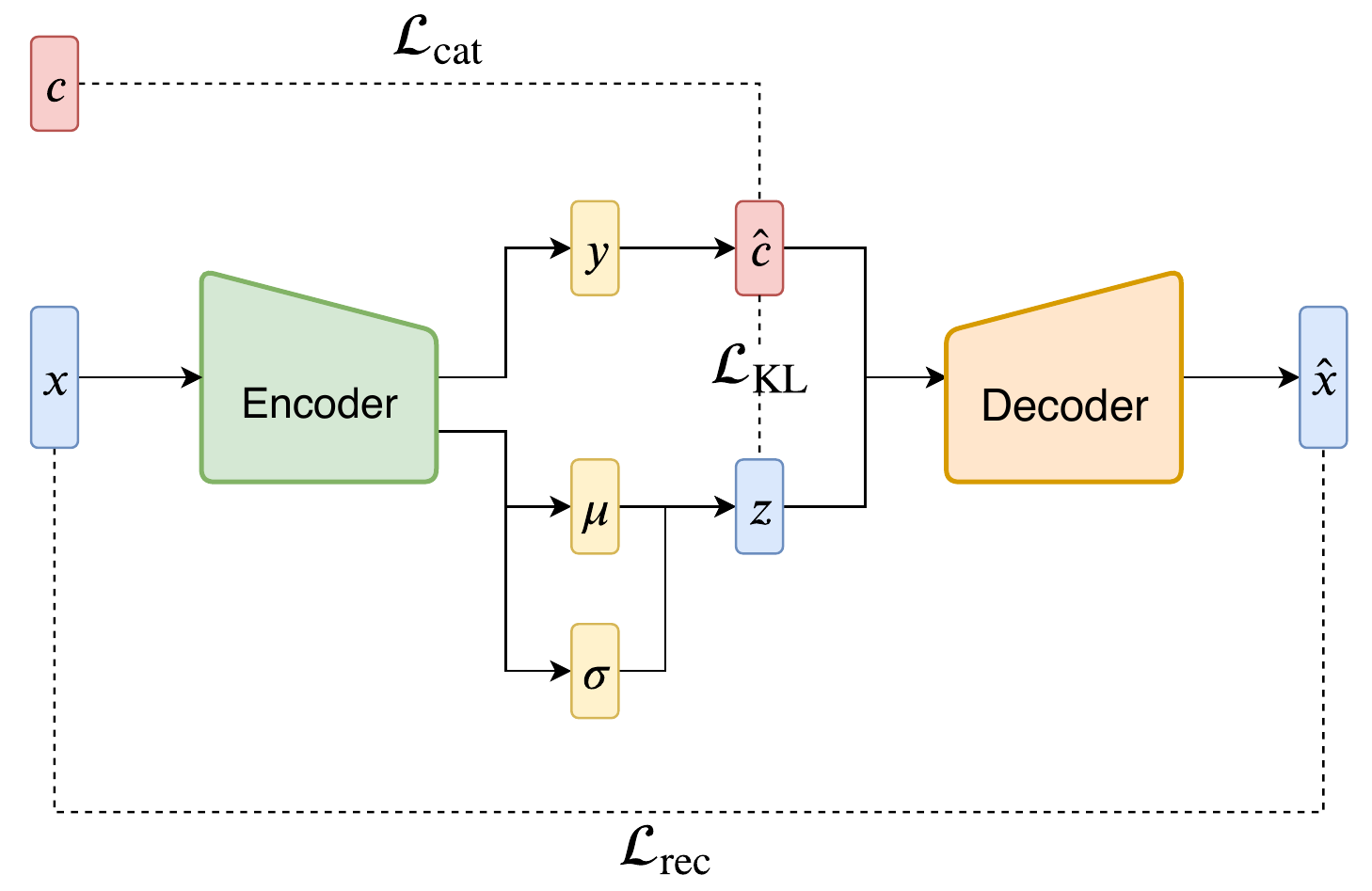}
    \caption{}
    \label{fig:model}
    \end{subfigure}
    \hfill
    \begin{subfigure}{0.40\linewidth}
    \includegraphics[width=1.0\textwidth]{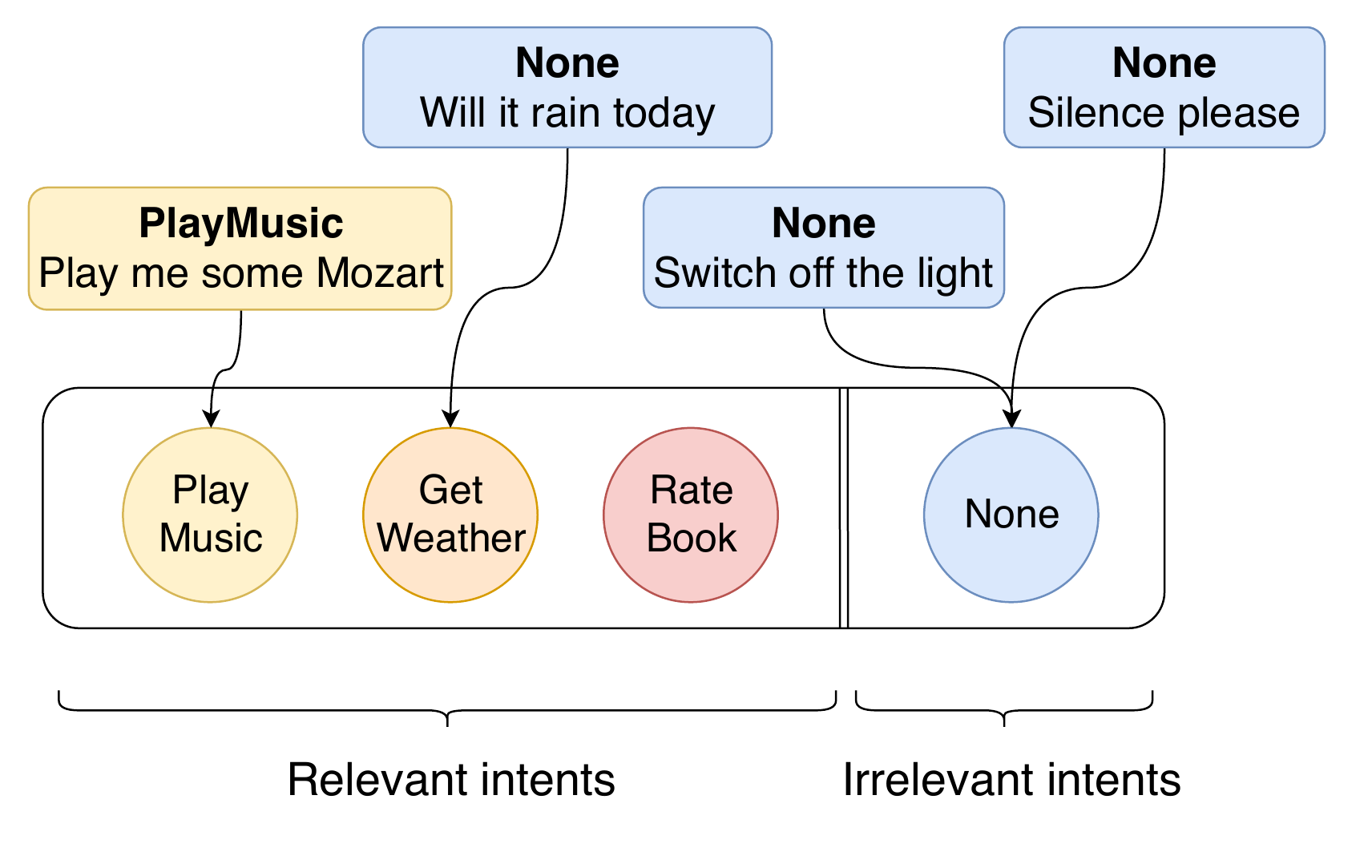}
    \caption{}
    \label{fig:categorical}
    \end{subfigure}
    \caption{Architecture of the model. (Left panel) The variational autoencoder architecture with the various losses defined in Eq.~\ref{eq:losses}. The code $\boldsymbol{z}$ is obtained from $(\boldsymbol{\sigma}, \boldsymbol{\mu})$ through the so-called {\it reparametrization trick} while the categorical variable is sampled using the {\it Gumbel trick} on the continuous vector $\boldsymbol{y}$. (Right panel) An illustration of the categorical latent vector and its role in filtering relevant sentences.}
\end{figure*}
	
    \subsection{Related work}
        \label{sec: rel_work}
        
        While there is a vast literature on text generation, conditional generation, data augmentation and transfer learning, there are only few existing works that combine these elements.
        In~\cite{kurata2016} and \cite{hou2018sequence} the authors use variational autoencoders to generate utterances through paraphrasing with the objective of augmenting the SLU training set and improve slot-filling. There is no conditioning on the intent and the data used to train the paraphrasing model is annotated and in-domain.
        
        In~\cite{yoo2019slot} the authors use a CVAE to generate queries conditioned to the presence of certain slots and observe improvements in slot-filling performance when augmenting the training set with generated data. In~\cite{yoo2019slu} they instead propose an autoencoder that is capable of jointly generating a query together with its annotation (intent and slots) and show improvements in both intent classification and slot-filling through data augmentation. In neither of the above, the model conditions the generation on the intent label nor leverages unannotated data for the training. 
        
        In a recent paper~\cite{cho2019}, the authors use semi-supervised self-learning to iteratively incorporate data coming from an unannotated set into the annotated training set. Their chosen metrics are both SLU performance and query diversity. This method represents a valid alternative to our generative data augmentation protocol and will be the object of competitive benchmarks in future work, where the impact of training data augmentation on SLU performance will be explored.

\section{Approach}
    \label{sec: approach}
    
    \textbf{Conditional variational autoencoders.} In order to generate queries conditioned to an underlying intent, we use a CVAE as depicted on~Fig.~\ref{fig:model}.
	While with VAEs the latent vector only incorporates continuous variables \cite{kingma2013auto}, features of discrete nature can be considered as input in CVAEs, e.g. the digit-class in MNIST or the user intent in conversational systems.
	Just like the encoding of continuous features, the encoding of discrete features is non-deterministic yet differentiable, thanks to the Gumbel-Max trick~\cite{maddison2016gumbel}, and regularized to match a simple prior, generally taken to be the uniform categorical distribution. 
	
	Each training sample is associated to a continuous feature vector $\boldsymbol{x}$ and a categorical variable expressed as a one-hot vector $\boldsymbol{c}$.
	Differently from~\cite{cvae_2015} and other implementations, we do not condition the generation of the latent code $\boldsymbol{z}$ to the categorical variable. We instead use the encoding distribution to generate the latent variable $\boldsymbol{\hat{c}}$ and we add both supervision and a KL regularization term that enforces the prior distribution on the classes.
	The associated loss function consists of three terms, namely the reconstruction term, the Kullback-Leibler term, and the categorical term:
	\begin{equation}
	\mathcal{L} = \mathcal{L}_{\rm rec} + \gamma \mathcal{L}_{\rm KL} + \mathcal{L}_{\rm cat},
	\label{eq:losses}
	\end{equation}
	where
	\begin{align*}
	    \mathcal{L}_{\rm rec} &= -\underset{q_{\phi}(\boldsymbol{z}| \boldsymbol{x})}{\mathbb{E}}[\log p_{\theta}(\boldsymbol{x} | \boldsymbol{z}, \boldsymbol{\hat{c}})] \\
	    \mathcal{L}_{\rm KL} &= D_{\mathrm{KL}}(q_{\phi}(\boldsymbol{z} | \boldsymbol{x}) \| p(\boldsymbol{z})) +D_{\mathrm{KL}}(q_{\phi}(\boldsymbol{c} | \boldsymbol{x}) \| p(\boldsymbol{c}))\\
	    \mathcal{L}_{\rm cat} &= -\sum_{i=1}^{C} c_i \, \alpha_{i} \,\log \left(q_{\phi}(c_i | \boldsymbol{x}) \right)
	\end{align*}
	In the equations above, $q_{\phi}$ and $p_{\theta}$ represent the encoder and the decoder respectively with their associated parameters, $C$ is the dimension of the categorical space.
	The constant $\gamma$ is used to set the relative weight of the KL regularization and perform annealing during training \cite{bowman2015generating, sonderby2016train}. We introduce the class-specific $\alpha$ coefficients to account for possible class-imbalance as well as to set the strength of the supervision we exercise on each category and. For all the experiments presented here, $p(\boldsymbol{c})$ is the uniform categorical distribution, and $p(\boldsymbol{z}) = \mathcal{N}(\vec 0,\mathds{1})$. At inference time, a sentence is generated by constructing the concatenation of a chosen $\boldsymbol{\hat{c}}$ with a sampled $\boldsymbol{z}$, feeding it to the decoder and extracting greedily the most probable sequence.
	
	\textbf{Query transfer.} A CVAE can be trained on a dataset of annotated queries, namely on $(\boldsymbol{x}, \boldsymbol{c})$ couples, where $\boldsymbol{x}$ is the sentence itself and $\boldsymbol{c}$ is the underlying query's intent.
	With too few sentences as training examples, a CVAE would not yield generated sentences of high enough quality and diversity. In addition to an annotated training dataset $\mathcal{D}_0$ -- kept small in the data scarcity regime of interest in this paper -- a second large ``reservoir'' dataset $\mathcal{D}_r$ is considered. The latter is unannotated and contains sentences that potentially cover a larger spectrum, ranging from intents that are semantically close to the in-domain ones to completely out-of-domain examples. 
	
% 	The queries from the reservoir are considered from a {\it None} intent, added to the categorical one-hot vector $\mathbf{c}$. Since we want to favor transfer of vocabulary, syntax or even entire queries from the reservoir to the target intents, we fix the coefficients in Eq. \ref{eq:losses} to be $\alpha_{\mathrm{c}}=1$ for all in-domain intents and $\alpha_{\mathrm{None}}=\alpha$, with $\alpha \in [0,1]$.
% 	We want to perform data augmentation for the original $\mathcal{D}_0$ by generating new sentences conditionally to the intents of $\mathcal{D}_0$.

    The novelty in our approach is that one extra dimension is allocated for irrelevant sentences coming from $\mathcal{D}_r$, namely an additional \textit{None} intent -- the categorical latent space of the CVAE already contains one dimension for each intent in $\mathcal{D}_0$. All sentences from $\mathcal{D}_r$ are supervised by a cross-entropy loss to this dimension, but we want the relevant ones to be allowed to transfer to one of the intents of $\mathcal{D}_0$, as illustrated on Fig.~\ref{fig:categorical}. To allow for this to happen, we may control the amount of transfer by weakening the supervision loss of $\mathcal{D}_r$ by the factor $\alpha$ (for simplicity, we will denote $\alpha_{\mathrm{None}}=\alpha$ in the following, namely $\forall i \neq \mathrm{None},  \alpha_i = 1 $). In the case $\alpha = 0$, the sentences from $\mathcal{D}_r$ are not supervised at all. The validity of this approach and the effect of $\alpha$ is illustrated in the context of computer vision in the Appendix.

    \textbf{Sentence selection.} We introduce another mechanism to further improve the query transfer. Since $\mathcal{D}_r$ may contain a lot of irrelevant data that can potentially pollute the generation and conditioning, we may want to preprocess it and select queries that belong to a close domain. In the context of natural language processing, this may be achieved by sentence embeddings. We suggest to use generalist sentence embeddings such as InferSent \cite{conneau2017supervised} as a first, rough, sentence selection mechanism. We first compute an ``intent embedding'' $\Vec{I}$ for each intent of $\mathcal{D}_0$, obtained by averaging the embeddings $\Vec{S}$ of all the sentences of the given intent. Then we only collect the sentences from $\mathcal{D}_r$ which are ``close'' enough to one of the intents of $\mathcal{D}_0$, i.e
    \begin{equation}
        \left\lbrace \Vec{S} \mid \exists \Vec{I}, \cos \left(\theta(\Vec{I}, \Vec{S})\right) > \beta \right\rbrace,
    \end{equation}
    where $\beta$ is a threshold which controls selectivity.

\begin{figure*}[htb]
    \centering
    \includegraphics[width=1.\textwidth]{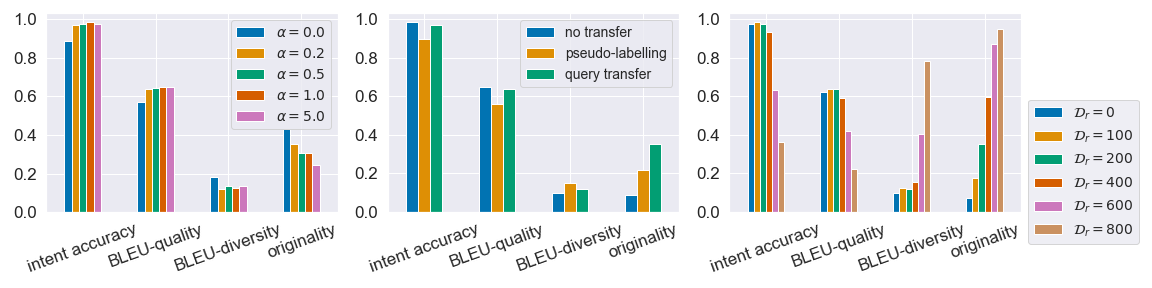}
    \caption{Generation metrics. (Left Panel) Evolution of the generation metrics as a function of the transfer parameter $\alpha$ for 200 sentences in both $\mathcal{D}_0$ and  $\mathcal{D}_r$. (Middle Panel) Comparison of the introduced method (transfer with InferSent selection at $\alpha=0.2$) with two baselines: one without any transfer ($|\mathcal{D}_r|=0$) and one with InferSent pseudo-labelling (see text). (Right Panel) Effect of the size of the reservoir $\mathcal{D}_r$ (for $|{D}_0|=200$ and $\alpha=0.2$) on the generation: increasing the number of transferred sentences improves the generation up to a certain point at which the quality degrades rapidly.}
    \label{fig:results}
\end{figure*}

\section{Experiments}
\label{sec: experiments}
    
    \subsection{Experimental setup}
    
        \textbf{Data processing.} For our experiments we use the publicly-available Snips benchmark dataset \cite{coucke2018snips}, which contains user queries from 7 various intents such as manipulating playlists or booking restaurants and 2000 queries per intent (from which we will only keep small fractions to mimic scarcity) and a test set of 100 queries per intent. Each intent comes with specific slots. As a proxy for a reservoir dataset, we use a large in-house dataset which collects assistants created by Snips users and contains all sorts of queries from over 300 varied intents.
        
    	The word embeddings feeded to the encoder are pre-trained GloVe embeddings~\cite{pennington2014glove}. We use a delexicalization procedure similar to that used in ~\cite{hou2018sequence} for Seq2Seq models.  First, slot values are replaced a placeholder and stored in a dictionary (``Weather in Paris'' $\rightarrow$ ``Weather in [City]''). The model is then trained on these delexicalized sentences and new delexicalized sentences are generated. A last step may consist in relexicalizing the generated sentences: abstract slot names are replaced by stored slot values. The effort is indeed put on generating new contexts, rather than just shuffling slot values.
    	
    	Note that if the slots are too loosely defined, one would in principle have to pay attention to context to relexicalize \cite{hou2018sequence}. Here we assume that the slots are sufficiently specific to ignore this issue. We tried various strategies for the initialization of slot-embeddings (e.g. the average of all slot values) and found that it had no impact in our experiments. We therefore initialize them with random embeddings.

    	\textbf{Training details.} Both the encoder and the decoder of our model use one-layer GRUs, with a hidden layer of size $256$ and both the continuous and categorical latent spaces are of size $8$ (for the categorical one: 7 intents + one \textit{None} class). %Details of the hyper-parameters used for the runs may be found in~\ref{tab:model}. 
    	We adopt the KL-annealing trick from \cite{bowman2015generating} to avoid posterior collapse: the weight of the KL loss term is annealed from $0$ to $1$ using the logistic function, at a time and a rate given by two hyper-parameters $t_{KL}$ and $r_{KL}$.
        The hyper-parameters were chosen to ensure satisfactory intent conditioning: $t_{KL}=300$ and $r_{KL}=0.01$.
        %, but were not optimized in any particular way since model selection is not particularly meaningful outside of a specific task.
        
        The {\it Adam} optimization method is used and we train for $50$ epochs at a learning rate of $0.01$ with a batch size of $128$. Depending on the size of $\mathcal{D}_0$, it takes a few dozens of minutes per experiment on a laptop. No word or embedding dropout is applied. The InferSent threshold is set to $\beta=0.9$. Note that we draw a fixed number of samples from both $\mathcal{D}_0$ and $\mathcal{D}_r$, however since we are in a data scarcity regime and only consider small $\mathcal{D}_0$, this draw entails high variability. Hence all results presented are averaged over five random seeds.

    \subsection{Generation metrics}
    
        Choosing relevant metrics for generation tasks is always a tricky yet interesting question. Generally speaking, one must optimize a trade-off between quality and diversity of the generated sentences. Indeed, for data augmentation purposes, we want the generated sentences to both be consistent with the original dataset and bring novelty, which is somewhat in contradiction. We use the following metrics to assess quality and diversity.
        
        To account for quality we first consider the \textbf{intent conditioning} accuracy. The generated sentences  need to be well-conditioned to the intent imposed in the one-hot categorical variable during generation. We train an intent classifier based on a logistic regression on the full Snips dataset (2000 queries per intent), reaching near-perfect accuracy on the test set. We use this ``oracle'' classifier as a proxy for evaluating the accuracy of the intent conditioning.
        We then assess the semantic quality of the generated sentences by considering what is referred to in the following as the \textbf{BLEU-quality}, namely the forward Perplexity \cite{zhao2017adversarially}, or the BLEU score \cite{papineni2002bleu} computed against the reference sentences of the given intent.
        
        %This can be measured by an "oracle" classifier reaching near-perfect test accuracy on a large training set. We feed it with a generated sentence and check that the intent predicted by the classifier matches the one we imposed for generation.\\
        
        To account for diversity, we consider the so-called \textbf{BLEU-diversity} defined as $1-\textrm{self-BLEU}$ where $\textrm{self-BLEU}$ is merely the BLEU score of the generated sentences of a given intent against the other generated sentences of the same intent \cite{zhu2018texygen}. Finally, the second diversity metric is what we call the \textbf{originality}. Indeed, enforcing diversity does not ensure that we are not just reproducing the training set. If the latter has high diversity, we may obtain high diversity by plagiarizing it. Therefore the originality is defined as the fraction of generated delexicalized queries that are not present in the training set.
        
        These four metrics take values in $\left[0 , 1\right]$. The three last metrics (BLEU, BLEU-diversity, originality) are evaluated intent-wise, which may be problematic if the intent conditioning of the generated sentences is poor. For example, if we condition to ``PlayMusic'' and the generated sentence is ``What is the weather ?'', the diversity metrics of the ``PlayMusic'' intent would be over-estimated while the quality would be under-estimated. To reduce this effect as much as possible, the computation of these metrics is therefore restricted to generated sentences for which the oracle classifier agrees with the conditioning intent. 

\section{Results}
\label{sec: results}
    The code to reproduce all of our experiments is publicly available on GitHub\footnote{ \url{https://github.com/snipsco/automatic-data-generation}}.
    
    \subsection{Quality of generation}
    
        For all of the experiments described in this paragraph, we set $|\mathcal{D}_0|=200$ and $\beta=0.9$. As stated in Section~\ref{sec: approach}, the $\alpha$ parameter allows to control the amount of transfer between $\mathcal{D}_0$ and the reservoir $\mathcal{D}_r$. The left panel of Fig.~\ref{fig:results} indeed shows that $\alpha$ is a useful cursor for the diversty-quality tradeoff. Increasing $\alpha$ yields generated sentences of higher quality (both in terms of intent conditioning and BLEU-quality) but lower diversity (in terms of BLEU-diversity and originality). Again, the optimal value of $\alpha$ is task dependent and needs to be tuned accordingly: some tasks would rather require high quality, others would require high diversity (see the Appendix for an illustration on images). 
        
        To test the efficiency of the query transfer, we compare it to two baselines. The first one is simply a CVAE trained only on $\mathcal{D}_0$ (in blue on the middle panel of Fig.~\ref{fig:results}). The second one, referred to as \textit{pseudo-labelling} (in orange on the figure), leverages queries from $\mathcal{D}_r$ directly associated to intents of $\mathcal{D}_0$ using InferSent-based similarity scores (the CVAE is trained without a \textit{None} class). If the $\beta$ parameter defined in Section~\ref{sec: approach} exceeds a certain threshold for a given intent, the sentence from $\mathcal{D}_r$ is directly added to the corresponding intent in $\mathcal{D}_0$, on which the CVAE is trained. The middle panel of Fig.~\ref{fig:results} shows that the proposed query transfer method improves the diversity metrics (especially the originality) of the generated sentences, with hardly any deterioration in quality. In comparison, the pseudo-labelling approach deteriorates significantly the quality of generated sentences.

        % and CVAE trained using queries injected from $\mathcal{D}_r$ by InferSent \textit{pseudo-labelling}. \\
        % \textbf{Infersent pseudo-labelling}: for every query in the reservoir, through InferSent embedding we can determine which one is the closest intent and, therefore, the query's pseudo-label. We can accept or not this pseudo-label based on the similarity score and the threshold $\beta$. We finally use queries from the reservoir and their pseudo-label during training of the CVAE
        % In all the experiments we present here we fixed the number of sentences in $\mathcal{D}_0$ to be 200 (a bit less than 30 sentences per intent), and selected a variable number of sentences from $\mathcal{D}_r$.
    
        Finally, the right panel of Fig.~\ref{fig:results} displays the evolution of the generation metrics with the size of the reservoir. We observe a remarkable improvement of the diversity metrics when the number of sentences injected from $\mathcal{D}_r$ increases, without any loss in quality up to a certain point at which the quality degrades strongly. This is due to the imbalance introduced in the conditioning mechanism of the CVAE. A statisfying trade-off seems to be found for $|\mathcal{D}_r|=|\mathcal{D}_0|$.

    \subsection{Data augmentation for language models}
        
        \begin{table}
         \setlength\tabcolsep{8pt}
            \centering
            \begin{tabular}{cccc}
            \hline
            $|D_0|$ & \begin{tabular}[c]{@{}c@{}} augmentation \\ ratio \end{tabular} & 
            \begin{tabular}[c]{@{}c@{}}PPL \\  $\mathcal{D}_{\mathrm{aug}}$\end{tabular}    &
            \begin{tabular}[c]{@{}c@{}}PPL \\  $\mathcal{D}_{\mathrm{ref}}$\end{tabular}    \\ \hline
            \multirow{2}{*}{$125$}  & $+50\%$                                                        & $-2.322$ & $-17.73$ \\ %\cline{2-5} 
                                  & $+100\%$                                                 
             & $-5.909$ & $-28.62$  \\ \hline
            \multirow{2}{*}{$250$}  & $+50\%$                                                       & $-1.756$ & $-17.72$  \\ %\cline{2-5} 
                                  & $+100\%$                   
            & $-3.755$  & $-22.85$  \\ \hline
            \multirow{2}{*}{$500$}  & $+50\%$                                                      & $-3.335$  & $-12.34$  \\ %\cline{2-5} 
                                  & $+100\%$              
            & $-4.046$  & $-18.55$  \\ \hline
            \multirow{2}{*}{$1000$} & $+50\%$                                                     & $-1.031$  & $-9.278$  \\ %\cline{2-5} 
                                  & $+100\%$                                                
            & $-0.511$  & $-13.62$  \\ \hline
            \end{tabular}
            \caption{\label{tab:perplexity} Relative loss of perplexity (\%) with respect to LM trained on the original dataset $\mathcal{D}_0$, when varying the size of $\mathcal{D}_0$ and the augmentation ratio. Perplexities -- averaged over 3 experiments -- are computed on the test set $\mathcal{D}_{\mathrm{test}}$ for LMs trained on $\mathcal{D}_0$, $\mathcal{D}_{\mathrm{aug}}$, and $\mathcal{D}_{\mathrm{ref}}$ respectively, when varying the size of $\mathcal{D}_0$ and the augmentation ratio.
            %$\mathcal{D}_0$ contains \textit{\#queries} sentences and, $\mathcal{D}_{\mathrm{aug}}$ and $\mathcal{D}_{\mathrm{ref}}$ have \textit{\#queries} + \textit{\#queries added} utterances. 
            Results can only be compared row-wise because of the vocabulary restriction (see text).}
        \end{table}
      
        In this section, we show that the proposed approach can effectively be used as data augmentation technique for language modeling tasks. Indeed, leveraging in-domain language models in cascaded approaches -- trained for a specific use case rather than in a large vocabulary setting -- allows to both reduce their size and increase their in-domain accuracy~\cite{SNIPS}. We hence propose to compare the perplexity \cite{bahl1983maximum} of Language Models (LM) trained on three datasets: the initial dataset of delexicalized sentences $\mathcal{D}_0$, $\mathcal{D}_{\mathrm{aug}}$ containing $\mathcal{D}_0$ augmented by sentences generated by the CVAE model trained on $\mathcal{D}_0$ with query transfer, and $\mathcal{D}_{\mathrm{ref}}$ containing  $\mathcal{D}_0$ augmented by ``real'' sentences from the original Snips benchmark dataset. 
    
        %We also dispose of $\mathcal{D}_{\mathrm{test}}$ containing the test sentences. $\mathcal{D}_{\mathrm{aug}}$ and $\mathcal{D}_{\mathrm{ref}}$ contain the same number of sentences.
        
        The SRILM toolkit \cite{stolcke2002srilm} was used to train 4-grams LMs with Kneser-Ney Smoothing \cite{kneser1995improved}. Perplexity is only comparable if the vocabulary supported by the various models is the same. To fix this issue, the words contained in at least $\mathcal{D}_{0}$, $\mathcal{D}_{\mathrm{aug}}$ and $\mathcal{D}_{\mathrm{ref}}$ are added as unigrams with a count $1$ in every LM. Finally, the CVAE might generate sentences already present in $\mathcal{D}_{0}$ but every sentence is kept only once. The perplexity is evaluated on a pool of 700 test sentences, for 4 different data regimes (i.e. sizes of $|\mathcal{D}_0|$). The experiment is repeated 3 times.  In this experiment, we set $\alpha=0.2$, $\beta=0.9$ and $|\mathcal{D}_r|=|\mathcal{D}_0|$, consistently with the previous section.
        
        Table~\ref{tab:perplexity} shows the results when varying the size of $\mathcal{D}_{0}$ and the number of sentences generated by the augmentation process (augmentation ratios of $50\%$ and $100\%$). We see that the perplexity on the test set is consistently lower when the LM is trained on $\mathcal{D}_{\mathrm{aug}}$ rather than when trained on $\mathcal{D}_{0}$, though it does not reach the performance of augmentation with real data ($\mathcal{D}_{\mathrm{ref}}$). The improvement is less significant as the dataset size increases, illustrating that most phrasings of the various intents are already covered in this data regime. These results are encouraging and show that this technique could be used as a data augmentation process for SLU tasks, especially in the low data regime.     
	
\section{Discussion and conclusion}
\label{sec: conclusions}
    We introduce a method to alleviate data scarcity in conditional generation tasks where one has access to a large unlabelled dataset containing some potentially useful information, using conditional variational autoencoders. We present this approach in the context of sentence generation, but the same can be applied e.g. to visual data as shown in the Appendix. We choose to focus on the low data regime, as it is the most relevant for user-customized closed-domain dialogue systems, where gathering manually annotated datasets is very cumbersome.
    
    Transferring knowledge from the large reservoir dataset $\mathcal{D}_r$ to the original dataset $\mathcal{D}_0$ comes with the risk of introducing unwanted information which may corrupt the generative model. However, this risk may be controlled by adjusting two parameters. First, we consider a selectivity threshold $\beta$ to adjust how much irrelevant data is discarded from $\mathcal{D}_r$ during preprocessing. The pre-processing procedure consists in evaluating the similarity of an example from $\mathcal{D}_r$ with the examples in $\mathcal{D}_0$ (in this context the cosine similarity of sentence embeddings). Second, we introduce a transfer parameter $\alpha$, adjusting the supervision of unlabelled examples from $\mathcal{D}_r$, low values of $\alpha$ facilitating transfer from the reservoir.
    
    %When $\alpha$ is high, examples from $\mathcal{D}_r$ are supervised to an extra ``None'' dimension of the categorical latent vector of the CVAE and there is little transfer.
    % To increase the amount of transfer from $\mathcal{D}_r$, one should lower $\alpha$. 
    
    In this paper, we mainly focus on assessing the performance of the proposed generation technique by both introducing \textit{quality} and \textit{diversity} metrics and show how the introduced parameters may help choosing the best trade-off. We also illustrate our approach on a small language modelling task. The full potentiality of this method for more complex SLU tasks still needs to be explored and will be the subject of a future work.

\section{Appendix}
\label{sec:appendix}
% REF TO APPENDIX IN INTRO, CCL, MAIN TEXT 
%\subsubsection*{Appendix A: Illustration on images}
	
		\begin{figure}[htb]
	    \centering
		\begin{subfigure}[b]{0.2\textwidth}
			\includegraphics[width=\textwidth]{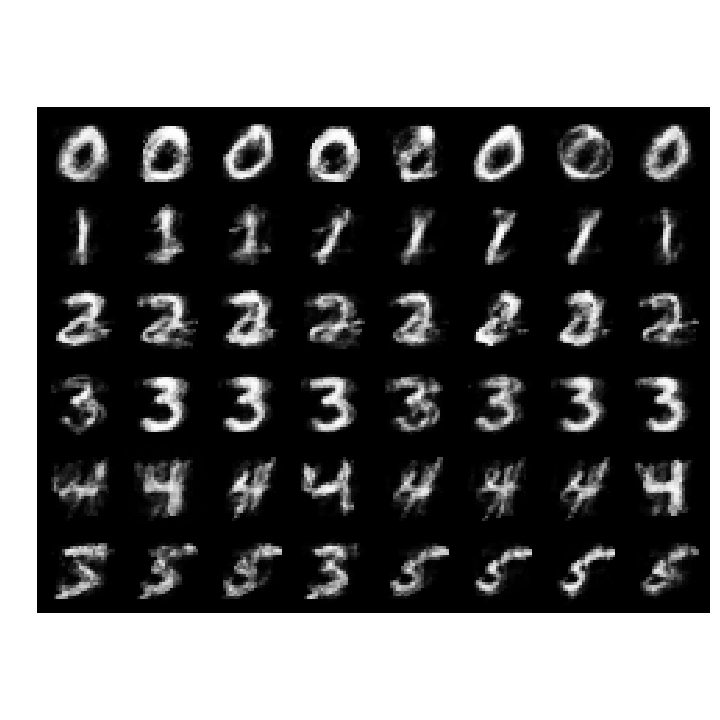}
			\vspace{-1cm}
			\caption{Without using $\mathcal{D}_r$}
		\end{subfigure}
		\begin{subfigure}[b]{0.2\textwidth}
			\includegraphics[width=\textwidth]{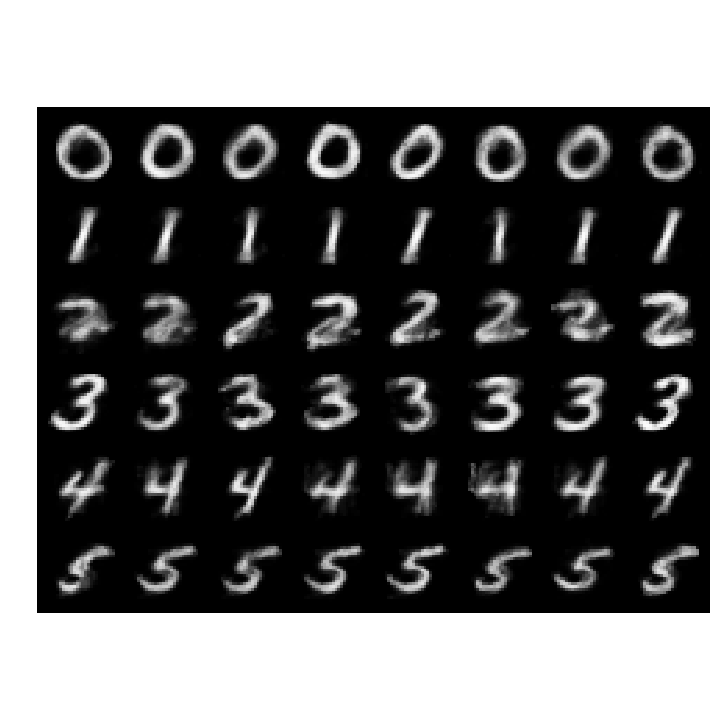}
			\vspace{-1cm}
			\caption{Using $\mathcal{D}_r$, $\alpha=0$}
		\end{subfigure}
		\begin{subfigure}[b]{0.2\textwidth}
			\includegraphics[width=\textwidth]{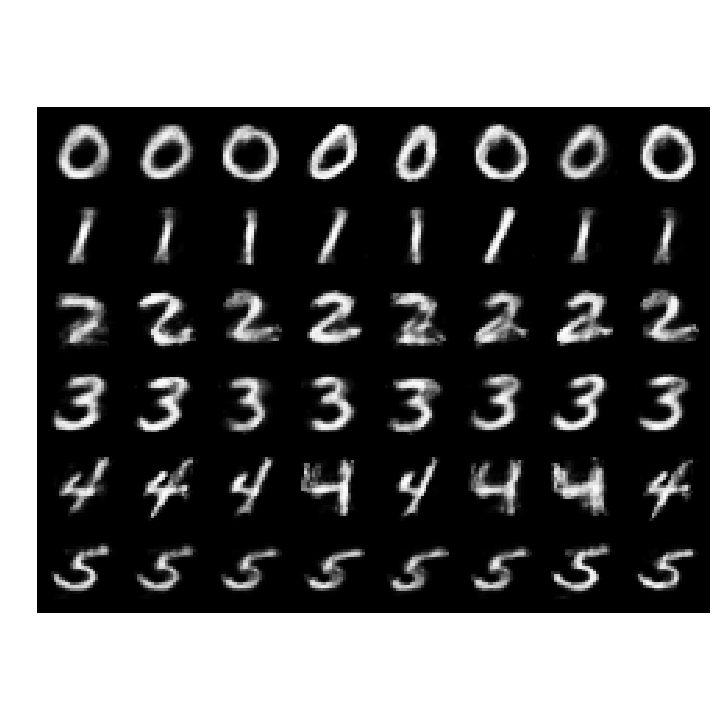}
			\vspace{-1cm}
			\caption{Using $\mathcal{D}_r$, $\alpha=1$}
		\end{subfigure}
		\begin{subfigure}[b]{0.2\textwidth}
			\includegraphics[width=\textwidth]{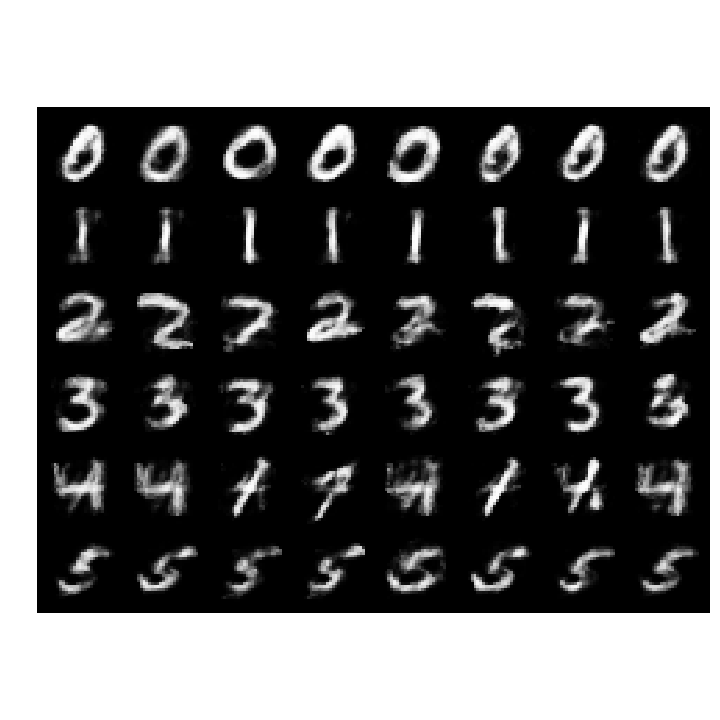}
			\vspace{-1cm}
			\caption{Using $\mathcal{D}_r$, $\alpha=2$}
		\end{subfigure}
		\caption{MNIST dataset. (a): CVAE trained on a small labelled dataset $\mathcal{D}_0$ of digits between 0 and 4 (10 images per class). (b)--(d): leveraging an unlabelled reservoir dataset $\mathcal{D}_r$ of digits between 0 and 9 (50 images per class), with a varying transfer parameter $\alpha$. Here the best quality/diversity trade-off is reached around $\alpha \sim 1$.}
		\label{fig:mnist}
	\end{figure}
	
	\begin{figure}[htb!]
	    \centering
        \begin{subfigure}[b]{0.2\textwidth}
			\includegraphics[width=\textwidth]{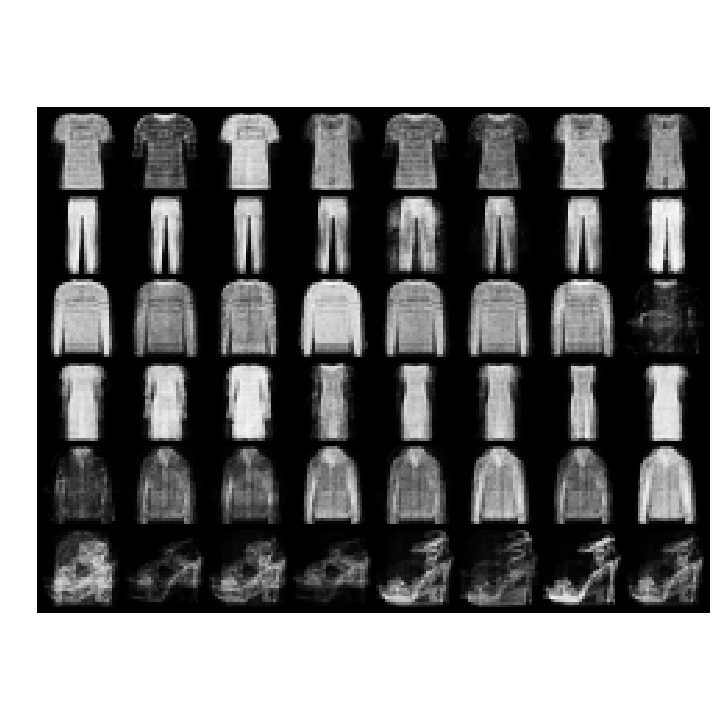}
			\vspace{-1cm}
			\caption{Without using $\mathcal{D}_r$}
		\end{subfigure}
		\begin{subfigure}[b]{0.2\textwidth}
			\includegraphics[width=\textwidth]{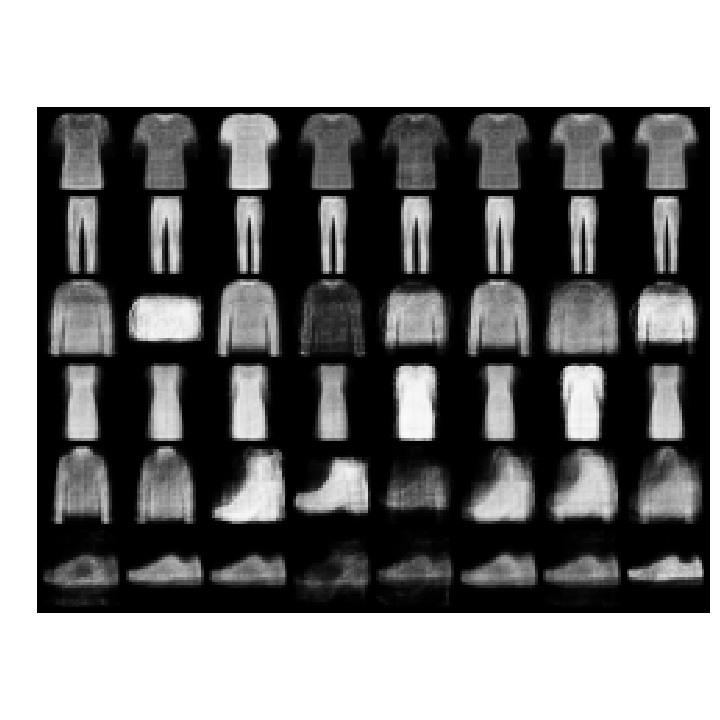}
			\vspace{-1cm}
			\caption{Using $\mathcal{D}_r$, $\alpha=0$}
		\end{subfigure}
		\begin{subfigure}[b]{0.2\textwidth}
			\includegraphics[width=\textwidth]{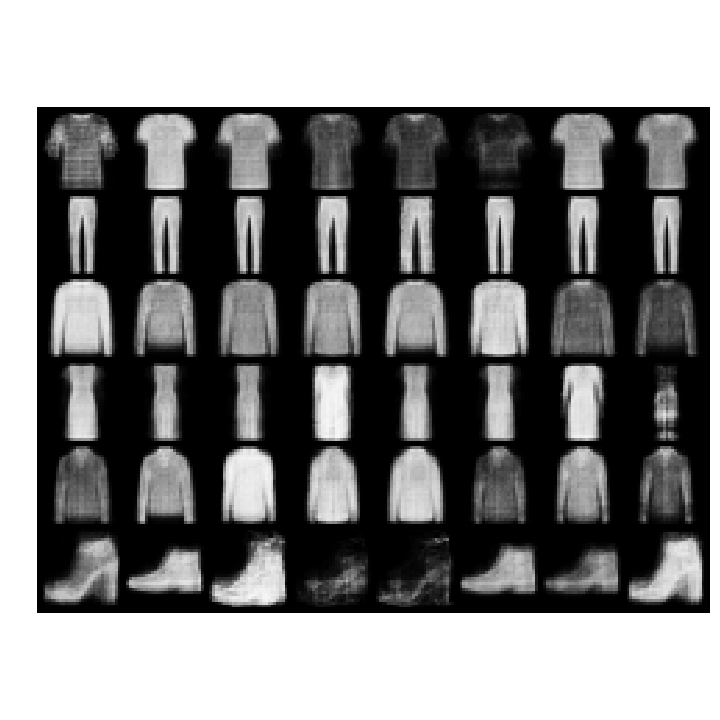}
			\vspace{-1cm}
			\caption{Using $\mathcal{D}_r$, $\alpha=0.1$}
		\end{subfigure}
		\begin{subfigure}[b]{0.2\textwidth}
			\includegraphics[width=\textwidth]{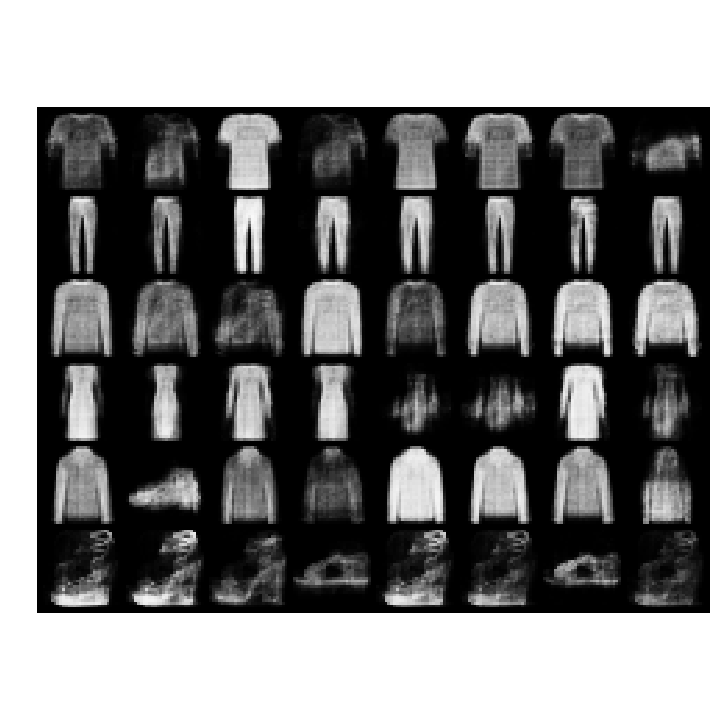}
			\vspace{-1cm}
			\caption{Using $\mathcal{D}_r$, $\alpha=1$}
		\end{subfigure}
		\caption{Fashion MNIST dataset. (a): CVAE trained on a small labelled dataset $\mathcal{D}_0$ containing only the first 5 classes (10 images per class). (b)--(d): leveraging an unlabelled reservoir dataset $\mathcal{D}_r$ containing all 10 classes (50 images per class), with a varying transfer parameter $\alpha$. Here the best quality/diversity trade-off is reached at $\alpha \sim 0.1$.}
		\label{fig:fashion_mnist}
	\end{figure}
	
	We present below results on the MNIST and Fashion MNIST \cite{xiao2017fashion} datasets as toy examples to give another illustration of the transfer process. Here, the small annotated dataset $\mathcal{D}_0$ contains only examples from the first 5 first classes of each dataset, with 10 examples per class. The larger reservoir dataset $\mathcal{D}_r$ contains examples from each of the 10 classes (half of its content being irrelevant to the generative task), with 50 examples per class.
	
	Figs. \ref{fig:mnist} \& \ref{fig:fashion_mnist} show results obtained by training a very simple two-layer fully-connected conditional variational auto encoder for 200 epochs, for various values of the transfer parameter $\alpha$. The code used to produce these figures is included in the GitHub repository. We see that without the reservoir dataset $\mathcal{D}_r$ (panel (a) on both figures), there is not enough training data to generate high-quality diverse images. Using $\mathcal{D}_r$ with a too low value of $\alpha$ (second column) yields unwanted image transfer and corruption of the generated images (4's get mixed up with 9's and 7's in MNIST, shoes get mixed up with jackets in Fashion MNIST). Conversely, if $\alpha$ is too high (panels (d) on both figures), there is not enough transfer and the generated images do not benefit from $\mathcal{D}_r$ anymore. However, for a well-chosen value $\alpha^\star$ (panels (c)), there is significant improvement both in quality and diversity of the generated images. As we can see here, the optimal value of $\alpha$ is dataset-dependent: $\alpha^\star\sim1$ for MNIST, $\alpha^\star\sim 0.1$ for Fashion MNIST.

%\vfill\pagebreak
\bibliographystyle{IEEE}
\bibliography{strings,acl2019}

\end{document}